\documentclass[review]{elsarticle}

\usepackage{lineno,hyperref}
\modulolinenumbers[5]
\usepackage{times}
\usepackage{epsfig}
\usepackage{graphicx}
\usepackage{amsmath}
\usepackage{amssymb}
\usepackage{array}
\usepackage{pgfplots,pgfplotstable}
\usepackage{tikz}
\usepackage{algorithm}% http://ctan.org/pkg/algorithm
\usepackage{algpseudocode}% http://ctan.org/pkg/algorithmicx

\journal{Applied Soft Computing}

\begin{document}

\begin{frontmatter}

\title{Novel Adaptive Genetic Algorithm Sample Consensus}

\author[mymainaddress,Ehsan]{Ehsan Shojaedini}
\author[mymainaddress,Mahshid]{Mahshid Majd}
\author[mymainaddress,Reza]{Reza Safabakhsh}
\address[mymainaddress]{Amirkabir University of Technology, Tehran, Iran\\}
\fntext[Ehsan]{shojaedini@aut.ac.ir}
\fntext[Mahshid]{majd@aut.ac.ir}
\fntext[Reza]{safa@aut.ac.ir}

\begin{abstract}
Random sample consensus (RANSAC) is a successful algorithm in model fitting applications. It is vital to have strong exploration phase when there are an enormous amount of outliers within the dataset. Achieving a proper model is guaranteed by pure exploration strategy of RANSAC. However, finding the optimum result requires exploitation. GASAC is an evolutionary paradigm to add exploitation capability to the algorithm. Although GASAC improves the results of RANSAC, it has a fixed strategy for balancing between exploration and exploitation.
In this paper, a new paradigm is proposed based on genetic algorithm with an adaptive strategy. We utilize an adaptive genetic operator to select high fitness individuals as parents and mutate low fitness ones. In the \textit{mutation} phase, a training method is used to gradually learn which gene is the best replacement for the mutated gene. The proposed method adaptively balance between exploration and exploitation by learning about genes. During the final Iterations, the algorithm draws on this information to improve the final results. 
The proposed method is extensively evaluated on two set of experiments. In all tests, our method outperformed the other methods in terms of both the number of inliers found and the speed of the algorithm.
\end{abstract}


\end{frontmatter}

\section{Introduction}

Regression in the presence of outliers is a challenging task. In real time tasks such as visual odometry and projective transformations, using a fast and reliable regression technique is vital. RANSAC\cite{fischler1981random} is one of the most successful regression techniques that is well applicable to outlier contaminated data. It follows a simple assumption, a small set of data that is not contaminated with outlier will construct a model with the lowest error.

RANSAC repeatedly performs two simple steps: hypothesis generation and evaluation. It generates the hypothesis randomly by selecting a small random set of points. Using each set, a model is built and assigned to its hypothesis. Next, each generated hypotheses are tested on all of the data points. Many researchers tried to improve RANSAC by refining these two steps. Some researchers tried to guide the algorithm by presenting extra information in order to decrease the randomness of generation step and increase the chance of generating better hypotheses. Other works focus on redesigning hypothesis evaluation to speed up the algorithm which is crucial in case of large datasets.

Mapping RANSAC to an evolutionary algorithm, it can be considered as a simple genetic algorithm with the population size of one and a \textit{mutation} operator that mutates the population in each iteration. The first attempt to add all genetic operators to RANSAC is GASAC \cite{rodehorst2006genetic} which guides the hypothesis generation through the solution path.
In contrast to RANSAC which purely explores the solution area, GASAC gives a 50\% chance to both exploration and exploitation. Although GASAC outperforms RANSAC, it suffers a major drawback. It blindly uses a fixed threshold for \textit{selection} operator. The threshold has a significant impact on the success of the algorithm while the best value could be different not only in various problems but also in different iterations of the algorithm. A fixed threshold might lead the algorithm to select improper parents for \textit{crossover} or lose suitable individuals in \textit{mutation}.

In this paper, a novel adaptive approach is proposed which is based on GASAC. The main contribution of this paper is first, presenting a new adaptive \textit{selection} operator for both \textit{crossover} and \textit{mutation}, designed for each individual and second, proposing a new \textit{mutation} operator which gradually learns the excellency of genes and chooses the best ones as replacements. The proposed algorithm successfully balances exploration against exploitation according to the outlier ratio in the dataset.

To evaluate our model, two sets of experiments are designed and reported. First, the algorithm is evaluated in the single feature matching problem where the inlier matches should be discriminated. The test is repeated for a hundred times to achieve a reliable conclusion. In the second sets of experiments, the algorithm is applied to solve the perspective-three-point problem \cite{gao2003complete} in a real-world stereo visual odometry dataset. In this problem, the camera pose is estimated using a set of 3D points and their corresponding 2D projection on the camera images. 
Both results are compared to the ones of RANSAC and GASAC and a significant improvement is reported.

The rest of the paper is organized as follows. The related literature is reviewed in Section 2. In Section 3 the evolutionary based RANSAC algorithms are presented. Section 4 presents our novel adaptive genetic algorithm sample consensus. Section 5 gives the experimental results followed by the conclusions given in Section 6.

\section{Related Work}

It starts with Fishler and Bolles paper which proposed Random Sample Consensus called RANSAC. This is a method for fitting a model to experimental data which contains a significant percentage of outliers. RANSAC repeatedly selects a minimum random subset of data and build a model based on them. Each model is a hypothesis, and the number of data points fitted in the model is counted. The final model will be constructed using all points that achieve a consensus on the best model. In other words, let the set of all points be $ P $ and the minimal size of the set to construct the hypothesis be $ n $. In each iteration, a subset $ S_i $ data points is selected randomly from $ P $ and construct the $ M_i $ model. The model is verified against all data points in $ P $ and a subset $ S^*_i $ containing all the points that support the hypothesis is determined, These points are called \textit{H-inliers} (Hypothesis inliers). These steps are repeated until the probability of finding a better model is lower than a specified threshold. The Final model is constructed using the points of the largest set of $ S^* $. 

RANSAC assumes that the outlier points does not have a consensus on a particular model. Therefore, it uses the \textit{H-inlier} count as criteria for finding the best model. Torr and Zisserman \cite{torr2000mlesac} proposed a maximum likelihood estimation by sampling consensus model called MLESAC. It is a new estimator that instead of just using number of \textit{H-inliers}, maximize the log likelihood of the solution using random sampling. MLESAC improves RANSAC using a better cost function in terms of the likelihood of inliers and outliers. Tordoff and Murray \cite{tordoff2002guided} explained that MLESAC has a major drawback, its speed. They extended MLESAC algorithm by guiding the selection of features which reduced the number of iterations required for a given confidence in the solution.

The speed of the algorithm depends on two factors: The duration of each iteration and the number of iterations. For the first factor, in each iteration there are two main processes: hypothesis generation, and hypothesis verification. As larger the size of the set of all points is, longer the hypothesis verification takes. A Randomized version of RANSAC (R-RANSAC) is introduced by Chum and Matas \cite{chum2002randomized} in order to improve the hypothesis verification efficiency. Their paradigm is to have a pre-test step before the verification on all points. Each hypothesis would be verified only if it passes the pre-test. In the pre-test step a randomized set of points is selected and verified by the hypothesis. If all points are verified as inlier then the complete verification is held. In this way, the useless models is detected and eliminated in the pre-test step. Therefore, many unnecessary verifications are eliminated and cause the R-RANSAC to be much faster than its ancestor RANSAC. 

Chum and Matas also proposed Locally Optimized RANSAC (LO-RANSAC) \cite{chum2003locally} to improve the hypothesis generation step of RANSAC. They generate a constant number of hypothesis using just the inlier set of the best current model. By doing so, the results of the algorithm improve more rapidly and termination criteria would met earlier. They completed the algorithm on their next papers \cite{chum2004enhancing,chum2008optimal} as LO-RANSAC-RD and Optimal Randomized RANSAC. 

In spite of the improvement on the speed of RANSAC, these algorithms are not still directly applicable in real-time application. Nister proposed Preemptive RANSAC \cite{nister2003preemptive} to cope with this problem. In this algorithm, first a fixed number of hypothesis is generated and half of the hypotheses eliminated in each iteration. Thus, the termination of the algorithm in a fixed time is granted. Tanaka \textit{et al.} modified preemptive RANSAC and proposed an incremental version of RANSAC \cite{tanaka2006incremental}. KALMANSAC \cite{vedaldi2005kalmansac} is another real-time approach which is used in tracking and the estimation of structure from motion (SFM). It can tolerate large number of outliers which is inspired from pseudo-Bayesian filtering algorithms. Adaptive Real-Time Random Sample Consensus
In (ARRSAC) \cite{raguram2008comparative} also set the number of hypothesis in the initial candidate to ensure that the runtime of the algorithm is bounded. ARRSAC is an adaptive version of preemptive RANSAC which operates in a partially depth-first manner.

Bail-out test \cite{capel2005effective} is another paradigm to terminate the verification early which is similar to R-RANSAC \cite{chum2002randomized}. More specifically, it considers the number of inliers $K_n$ within $C_n$ which is a subset of all points $P$ follows a hyper-geometric distribution $ K_n\sim HypG(K,n,\bar{K},N) $ ,where $ \bar{K} $ is number of inliers for $K_n$, $N$ total number of points in $P$, and $n$ size of $C_n$. It considers $ \bar{K}_{best} $ inlier count for the best current model and if the probability of $P(\bar{K}>\bar{K}_{best})$ is below a given threshold, the verification can continue.

Instead of generating the hypothesis randomly, PROSAC \cite{chum2005matching} draw samples from progressively larger sets of top-ranked correspondences. PROSAC uses a similarity function and select a subset of points with the highest similarity. The Progressive Sample Consensus (PROSAC) orders the points based on their similarity and progressively a larger set of points is used to generate new hypotheses. However, Michaelsen \textit{et al.} work called GOODSAC \cite{michaelsen2006estimating} uses a controlled search for selection of good samples. 

Purposive Sample Consensus (PURSAC) \cite{wang2015purposive} is similar to LO-RANSAC and instead of selecting samples from all of the available data, next sample subset is selected only form the best model inliers. In order to dilute the effect of sampling noise, PURSAC uses some different strategies such as selecting subset from data points that have long distance or using all of the inliers to generate a model.

jia \textit{\textit{et al.}} \cite{jia2016novel} proposed a new improved probability guide RANSAC algorithm (IPGSAC) which judges each data point based on its probability of being an inlier or outlier point. IPGSAC assumes that all of the data points follow a hybrid distribution shown in Eq. (\ref{eq:hypdis}) where $P(P_i)$ is the probability of estimated model $p_i$ and $e_i$ is its error and $c$ is expectectaion of inlier error and the probability of outlier point is $v$ and $\gamma$ determine $p_i$ is outlier or inlier. Each probability is updated using Eq. (\ref{eq:probUp}) where $M$ contain all of the \textit{H-inliers} points and $s$ is the estimated model. jia \textit{et al.} use DS evidence theory \cite{shafer1976mathematical} to achieve a more robust evaluation of test points.It starts with Fishler and Bolles paper which proposed Random Sample Consensus called RANSAC. This is a method for fitting a model to experimental data which contains a significant percentage of outliers. RANSAC repeatedly selects a minimum random subset of data and build a model based on them. Each model is a hypothesis, and the number of data points fitted in the model is counted. The final model will be constructed using all points that achieve a consensus on the best model. In other words, let the set of all points be $ P $ and the minimal size of the set to construct the hypothesis be $ n $. In each iteration, a subset $ S_i $ data points is selected randomly from $ P $ and construct the $ M_i $ model. The model is verified against all data points in $ P $ and a subset $ S^*_i $ containing all the points that support the hypothesis is determined, These points are called \textit{H-inliers} (Hypothesis inliers). These steps are repeated until the probability of finding a better model is lower than a specified threshold. The Final model is constructed using the points of the largest set of $ S^* $. 

RANSAC assumes that the outlier points does not have a consensus on a particular model. Therefore, it uses the \textit{H-inlier} count as criteria for finding the best model. Torr and Zisserman \cite{torr2000mlesac} proposed a maximum likelihood estimation by sampling consensus model called MLESAC. It is a new estimator that instead of just using a number of \textit{H-inliers}, maximize the log likelihood of the solution using random sampling. MLESAC improves RANSAC using a better cost function in terms of the likelihood of inliers and outliers. Tordoff and Murray \cite{tordoff2002guided} explained that MLESAC has a major drawback, its speed. They extended MLESAC algorithm by guiding the selection of features which reduced the number of iterations required for a given confidence in the solution.

The speed of the algorithm depends on two factors: The duration of each iteration and the number of iterations. For the first factor, in each iteration, there are two main processes: hypothesis generation, and hypothesis verification. As larger the size of the set of all points is, longer the hypothesis verification takes. A Randomized version of RANSAC (R-RANSAC) is introduced by Chum and Matas \cite{chum2002randomized} in order to improve the hypothesis verification efficiency. Their paradigm is to have a pre-test step before the verification on all points. Each hypothesis would be verified only if it passes the pre-test. In the pre-test step, a randomized set of points is selected and verified by the hypothesis. If all points are verified as inlier then the complete verification is held. In this way, the useless models are detected and eliminated in the pre-test step. Therefore, many unnecessary verifications are eliminated and cause the R-RANSAC to be much faster than its ancestor RANSAC. 

Chum and Matas also proposed Locally Optimized RANSAC (LO-RANSAC) \cite{chum2003locally} to improve the hypothesis generation step of RANSAC. They generate a constant number of hypothesis using just the inlier set of the best current model. By doing so, the results of the algorithm improve more rapidly and termination criteria would meet earlier. They completed the algorithm on their next papers \cite{chum2004enhancing,chum2008optimal} as LO-RANSAC-RD and Optimal Randomized RANSAC. 

In spite of the improvement in the speed of RANSAC, these algorithms are not still directly applicable in a real-time application. Nister proposed Preemptive RANSAC \cite{nister2003preemptive} to cope with this problem. In this algorithm, first a fixed number of hypotheses are generated and half of the hypotheses eliminated in each iteration. Thus, the termination of the algorithm in a fixed time is granted. Tanaka \textit{et al.} modified preemptive RANSAC and proposed an incremental version of RANSAC \cite{tanaka2006incremental}. KALMANSAC \cite{vedaldi2005kalmansac} is another real-time approach which is used in tracking and the estimation of structure from motion (SFM). It can tolerate a large number of outliers which is inspired by pseudo-Bayesian filtering algorithms. Adaptive Real-Time Random Sample Consensus
In (ARRSAC) \cite{raguram2008comparative} also set the number of hypothesis in the initial candidate to ensure that the runtime of the algorithm is bounded. ARRSAC is an adaptive version of preemptive RANSAC which operates in a partially depth-first manner.

Bail-out test \cite{capel2005effective} is another paradigm to terminate the verification early which is similar to R-RANSAC \cite{chum2002randomized}. More specifically, it considers the number of inliers $K_n$ within $C_n$ which is a subset of all points $P$ follows a hyper-geometric distribution $ K_n\sim HypG(K,n,\bar{K},N) $ ,where $ \bar{K} $ is number of inliers for $K_n$, $N$ total number of points in $P$, and $n$ size of $C_n$. It considers $ \bar{K}_{best} $ inlier count for the best current model and if the probability of $P(\bar{K}>\bar{K}_{best})$ is below a given threshold, the verification can continue.

Instead of generating the hypothesis randomly, PROSAC \cite{chum2005matching} draw samples from progressively larger sets of top-ranked correspondences. PROSAC uses a similarity function and selects a subset of points with the highest similarity. The Progressive Sample Consensus (PROSAC) orders the points based on their similarity and progressively a larger set of points is used to generate new hypotheses. However, Michaelsen \textit{et al.} work called GOODSAC \cite{michaelsen2006estimating} uses a controlled search for selection of good samples. 

Purposive Sample Consensus (PURSAC) \cite{wang2015purposive} is similar to LO-RANSAC and instead of selecting samples from all of the available data, next sample subset is selected only form the best model inliers. In order to dilute the effect of sampling noise, PURSAC uses some different strategies such as selecting a subset of data points that have long distance or using all of the inliers to generate a model.

jia \textit{\textit{et al.}} \cite{jia2016novel} proposed a new improved probability guide RANSAC algorithm (IPGSAC) which judges each data point based on its probability of being an inlier or outlier point. IPGSAC assumes that all of the data points follow a hybrid distribution shown in Eq. (\ref{eq:hypdis}) where $P(P_i)$ is the probability of estimated model $p_i$ and $e_i$ is its error and $c$ is expectation of inlier error and the probability of outlier point is $v$ and $\gamma$ determine $p_i$ is outlier or inlier. Each probability is updated using Eq. (\ref{eq:probUp}) where $M$ contain all of the \textit{H-inliers} points and $s$ is the estimated model. jia \textit{et al.} use DS evidence theory \cite{shafer1976mathematical} to achieve a more robust evaluation of test points.

\begin{equation} \label{eq:hypdis}
P(P_i)= \gamma \dfrac{1}{\sqrt{2 \pi } \sigma }exp\left(-\dfrac{(e_i - c)^2}{2 \sigma^2} \right) + (1-\gamma)\dfrac{1}{v},
\end{equation}
\begin{equation} \label{eq:probUp}
P(p_i \in M) = P(p_i \in M | s \not\subset M)P(s \not\subset M)+P(p_i \in M | s \subseteq M)P(s \subseteq M),
\end{equation}

\section{Evolutionary approaches of Sample Consensus}

One of the first usages of genetic algorithm in computer vision is presented by Saito and Mori \cite{saito1995application} paper in stereo matching. As the benefits of genetic algorithm have revealed, researchers utilized this technique in other contexts of computer vision as well \cite{chai1998robust,chai1998evolutionary,mingxing2002epipolar,hu2004robust}. Genetic Algorithm Sample Consensus (GASAC) \cite{rodehorst2006genetic} was one of the first forms of evolutionary approach in RANSAC algorithm. 

Briefly said, a simple genetic algorithm starts with a random population. Each individual of the population represents a solution for the problem. Based on the fitness of each answer, a subset of the population is selected as parents for the next generation. This step is called \textit{Selection}. A new generation is built combining parents genomes and this operation is called \textit{crossover}. Next, some parts of the genome would change randomly as \textit{mutation} to prevent the early convergence to a suboptimal solution. Finally, the population is sorted based on the fitness of individuals and a fixed number of them are selected as the next generation. These steps are repeated until a termination criteria is met.

RANSAC resembles a genetic algorithm with the population size of one where \textit{mutation} will perform on all parts of the chromosomes. Genetic Algorithm Sample Consensus (GASAC) improves this idea by increasing the population size and decreasing the \textit{mutation} probability. It also adds the \textit{crossover} operation to the algorithm. Each chromosome in GASAC contains a number of \textit{genes} equal to the minimum number of data points needed to solve the problem. Let a chromosome be $ G $, and the minimum number of data points needed to solve the problem be $ m $, then $ G=\{g_1,...,g_m\} $ and $ g_i\in\{1,...,n\} $ where $ n $ is the total number of data points. 

In each iteration, half of the population is selected as parents based on their fitness. The new generation is constructed applying \textit{crossover} operator on selected parents. Since the Randomness is a key point in the success of RANSAC, \textit{mutation} is an inevitable part of the algorithm. In this step each \textit{gene}, $ g_i$ would change to a random correspondence with the probability of $P_M=\dfrac{1}{2m} $.

Instead of blind randomness in RANSAC, GASAC tries to guide the population to have higher fitness while keeping a portion of randomness. The next generation is built upon the selected parents which have higher fitness and \textit{mutation} happens only by a defined chance. Although the provided guidance is simple, it shows to be productive with reported results. 

Vasconcelos \textit{et al.} \cite{vasconcelos2011adaptive} proposed an adaptive and hybrid version of GASAC. They proposed to change the \textit{mutation} probability when there is no improvement in a generation. Moreover, in order to improve the diversity of the population, if the best solution does not improve after a predefined number of generations, half of the population is replaced with a random generated individual based on their fitnesses. They also proposed a hybrid of GASAC and simulate annealing which improved the GASAC algorithm result. Another evolutionary approach of sample consensus is presented by Toda and Kubota \cite{toda2014behavior,toda2016evolution} as Evolution Strategy Sample Consensus (ESSAC). It uses a search range control method to reduce the search space and searching for the best solution is continued in the reduced space. In addition, the uses an adaptive \textit{mutation} probability which is updated in each generation based on the fitness of the best solution. Equation \ref{eq:essac} shows the adaptive \textit{mutation} probability of ESSAC where $fit_{best}$ is the fitness value of the best individual and $T_m$ is coefficient.

\begin{equation} \label{eq:essac}
P_m= exp\left(-\dfrac{fit_{best}}{T_m} \right)
\end{equation}

Otte \textit{et al.} \cite{otte2014antsac} proposed another evolutionary approach called ANTSAC. It is a new model based on Ant Colony Algorithms which uses the pheromone evaporation as a volatile memory. Each particular sample has a pheromone value which is updated in each iteration. The pheromone levels vaporize over time and ANTSAC chooses a particular sample based on the current pheromone matrix. 

\section{Novel Adaptive Genetic Algorithm Sample Consensus}

Genetic algorithms as a kind of heuristic optimization algorithm have two phases: \textit{exploration} and \textit{exploitation}. First, they explore the solution space for some good areas, and then exploit these areas for the best solution. When there is a large problem space and high rate of outliers, exploration phase is the cornerstone of success. Therefore, RANSAC which has only the exploration phase can cope with the high rate of outliers.
Still, without exploitation, RANSAC suffers a major drawback: it forgets suitable solutions and keeps only the best one. GASAC modifies RANSAC to alleviate this problem by generating new population from previous good individuals. The algorithm weakens the exploration phase by reducing \textit{mutation} probability and instead, adds exploitation using a \textit{crossover} operator. GASAC controls both exploration and exploitation phases with fixed thresholds throughout the generations. Yet, the optimum strategy would be a high rate of exploration at the beginning and gradually shift to exploitation.

Based on the above discussion, we propose a novel adaptive genetic algorithm sample consensus model which adaptively controls the exploration and exploitation phases. our contributions are in two folds:
\renewcommand{\labelitemi}{$\textendash$}
\begin{itemize}
	\item A new \textit{mutation} strategy is proposed where the excellency of genes are learned and the best ones are chosen as replacements. Consequently, \textit{mutation} would be fully random at early iterations and therefore the solution area is highly explored. Then it gradually learns the best replacements and carefully exploits the area.
	\item An adaptive selection mechanism is proposed to set the \textit{mutation} and \textit{crossover} probability of each individual based on their fitness. Note that the ratio of suitable parents to non-suitable ones could be different based on the dataset in use. Here, the fitnesses of individuals are judged locally according to the fitness of other individuals in each generation. Instead of a fixed threshold, the best and worst individuals are distinguished based on their normalized fitness values.
	
\end{itemize}

\subsection{Genetic Representation}
\begin{figure*}
	\begin{center}
		\includegraphics[width=.8\linewidth]{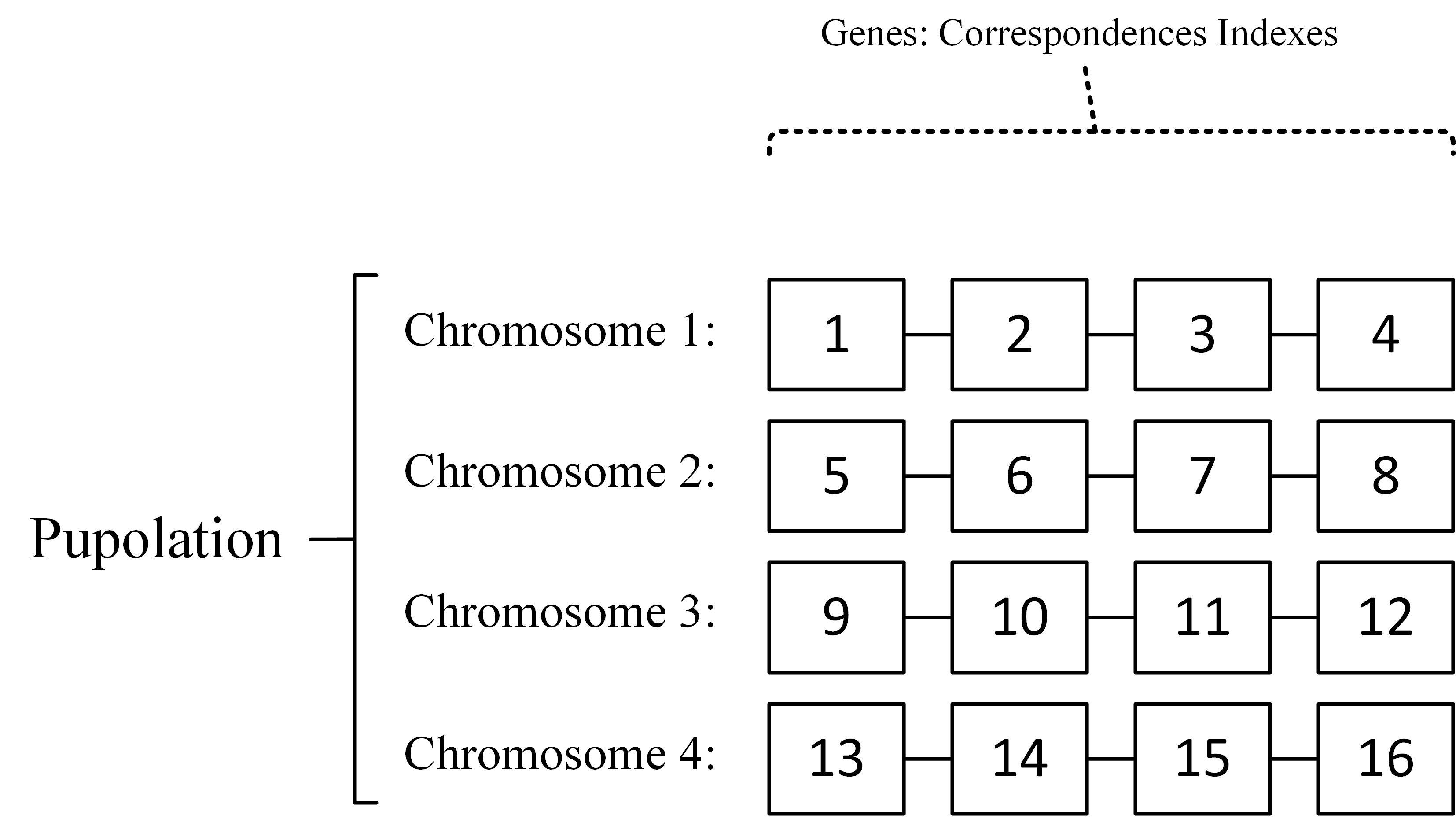}
	\end{center}
	\caption{Genetic representation of the perspective-three-point problem.}
	\label{fig:represent}
\end{figure*}

a chromosome represents a hypothesis for the optimum model. Like GASAC, each chromosome in our proposed method contains a minimal number of data point as its genes. As an example, to solve the perspective-three-point problem \cite{gao2003complete}, four correspondences are needed to estimate the camera motion. Therefore, each chromosome contains four genes. An example population of four chromosomes is represented in Figure \ref{fig:represent}.

\subsection{Fitness Function}

The fitness function is similar to the standard RANSAC evaluation method which is the number of \textit{H-inliers}. More specifically, for each chromosome a model of data is calculated using the genes of the chromosome, based on the task in hand. Then, the model is tested on the other data point and the number of the data point that supports the model are counted. In this way, the merit of each chromosome will be computed.

\subsection{Crossover and Mutation}

In order to generate the next generation population, we leverage two genetic operators: \textit{crossover} and \textit{mutation}. In our problem space, to have a chromosome with high fitness, all genes of the chromosome should be inlier. Therefore, changing one gene of a chromosome with a random gene affect the fitness significantly. Consequently, in this approach, \textit{mutation} performs as an exploration phase. However, when we have two chromosomes with high fitness, there is a high chance that all of the genes are inlier. So, changing the genes between these two chromosomes, would not change the fitness significantly. Thus, in this approach, \textit{crossover} acts as an exploitation phase.

\begin{figure*}
	\begin{center}
		\includegraphics[width=.8\linewidth]{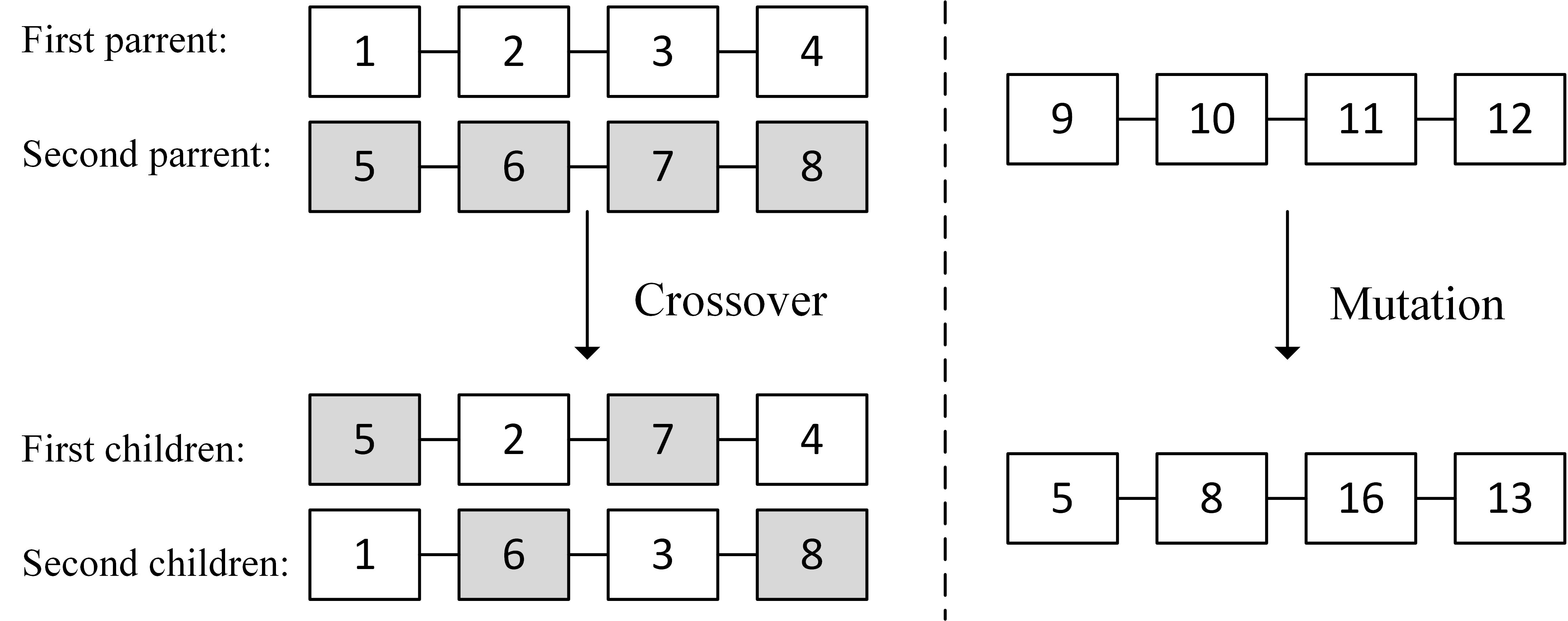}
	\end{center}
	\caption{An example of genetic operators.}
	\label{fig:operators}
	
\end{figure*}

In \textit{crossover} phase, two selected parents produce two children. Each child inherits its parent’s genes. we choose a fully random \textit{crossover} which means that each gene of the child is selected randomly from the first or second parent. An example of this operator is shown in Figure \ref{fig:operators}.

As mentioned before, \textit{mutation} is crucial to this approach as an exploration phase. The high rate of outliers necessitates a strong exploration. Therefore, if a chromosome is selected for mutation, all of its genes would change which is quite similar to the randomness of RANSAC. GASAC replaces the mutated genes with a random one. However, we take one step further and learn the best gene to replace with. 

\begin{figure*}
	\begin{center}
		\includegraphics[width=.8\linewidth]{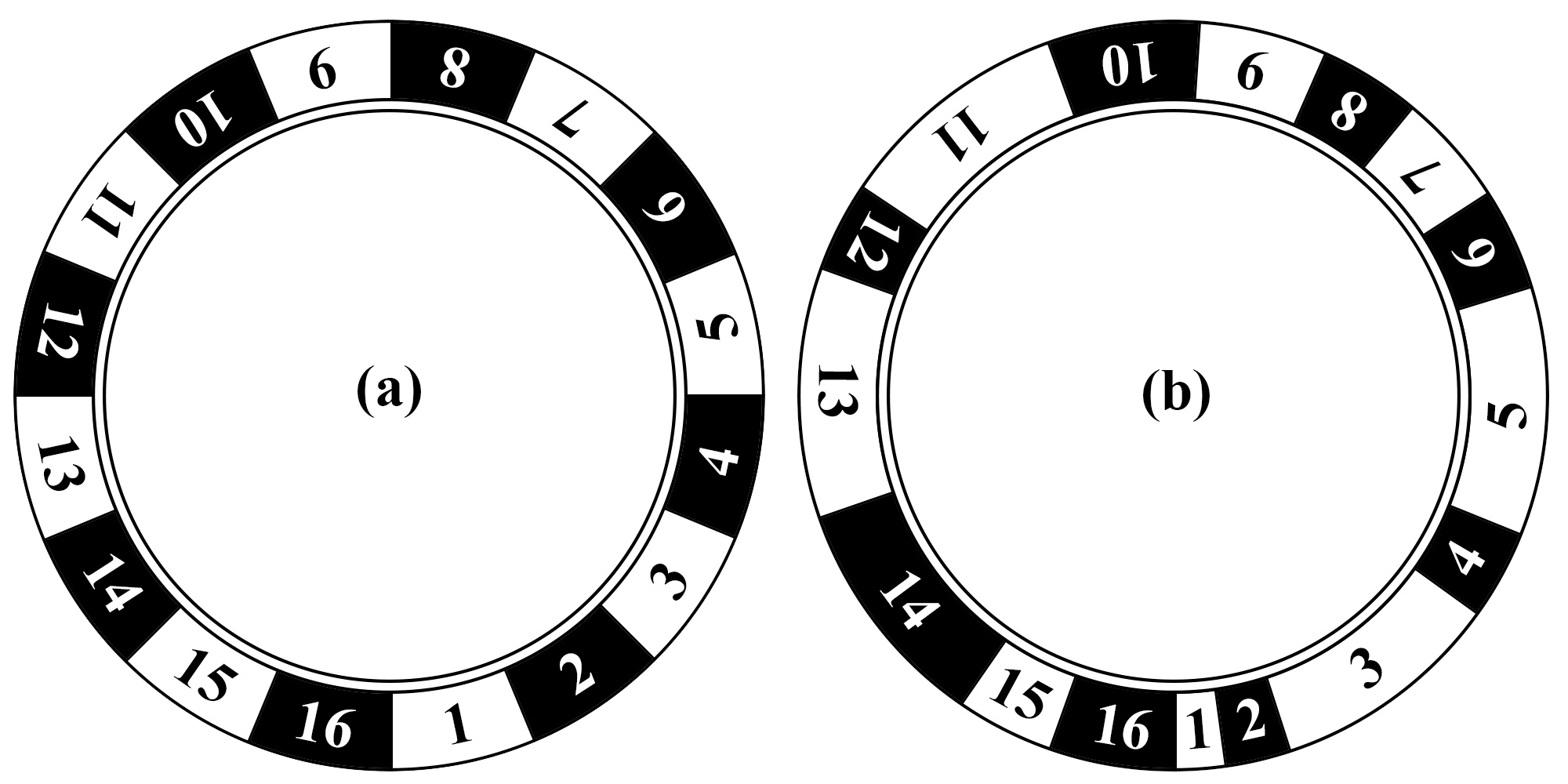}
	\end{center}
	\caption{(a) The roulette wheel at the start of the algorithm (b) The roulette wheel after training }
	\label{fig:roulettewheel}
	
\end{figure*}

To do so, we propose a technique to learn the probability of genes being inlier. As mentioned before, a chromosome with high fitness is likely to have inlier genes. We leverage this information to gradually learn which genes are more suitable. A roulette wheel is used to select the replacement genes in \textit{mutation} phase. Let the probability of selecting each gene to be $ P_g i $ which is initialized equally for all genes. In each iteration, the probability of genes is updated based on the fitness of the chromosomes they appear in. So that, the probability of genes in a chromosome with high fitness would increase and vice verse. Those genes that frequently appear in high fitness chromosomes would have higher selection probability. In other words, the algorithm learns which genes are suitable to construct a chromosome with high fitness. A sample of proposed roulette wheel before and after training is shown in Figure \ref{fig:roulettewheel}. Note that, this technique is quite different from fitness proportionate selection or roulette wheel selection because the probability of the geneses are learned gradually and the selection is not based on just the fitness of the chromosomes. 

\subsection{Selection}
RANSAC (as a genetic algorithm) selects all chromosomes for \textit{mutation} and GASAC takes the best half of chromosomes as parents. Both these algorithm would eventually find the optimum model. Nevertheless, It could be faster with an adaptive \textit{selection} operator. Towards a faster and more efficient algorithm, we propose an adaptive selection strategy for \textit{crossover} and \textit{mutation} to use each chromosome effectively. The goal is to only select the chromosomes that have a high fitness as parents in \textit{crossover} and those with low fitness to apply \textit{mutation}. Using a fixed threshold for selecting higher or lower fitness is not efficient since the range of fitnesses in each generation is different from other generations. Therefore, in the proposed method the fitnesses is normalized in each generation, then a power function is used to give a confidence in the selection of potential parent. The complete adaptive fitness based \textit{selection} operator equation for \textit{crossover} is formulated as follows:

\begin{equation} \label{eq:1}
Pc_i= norm\left(f\left(\textbf{g}_i\right)\right)^\gamma
\end{equation}
\begin{equation} \label{eq:2}
norm(f(\textbf{g}_i))=\dfrac{f\left(\textbf{g}_i\right)- \underset{j}{min}\{f(\textbf{g}_j)\} }{\underset{j}{max}\{f(\textbf{g}_j)\} - \underset{j}{min}\{f(\textbf{g}_j)\} }
\end{equation}

\begin{figure*}
	\begin{center}
		\resizebox{\columnwidth}{!}{%
			\begin{tabular}{c c}
				\begin{tikzpicture}
				\begin{axis}[
				minor y tick num=1,
				minor x tick num=1,
				extra y ticks={0},
				extra x ticks={0},
				extra tick style={grid style={black},xticklabel=\empty},
				grid=both,
				axis lines = left,
				xlabel = $f(\textbf{g}_i)$,
				ylabel = {$Pm_i$(\textit{mutation} probability of the $i$th genes)},
				]
				\addplot [
				domain=-0:1, 
				samples=100, 
				color=blue,
				]
				{exp(-x)}node[pos=0.25,anchor=west,fill=white,draw=black]{$\delta=1$};
				\addplot [
				domain=-0:1, 
				samples=100, 
				color=blue,
				]
				{exp(-x*2)}node[pos=0.28,anchor=west,fill=white,draw=black]{$\delta=\frac{1}{2}$};
				\addplot [
				domain=-0:1, 
				samples=100, 
				color=blue,
				]
				{exp(-x*5)}node[pos=0.350,anchor=west,fill=white,draw=black]{$\delta=\frac{1}{5}$};
				\addplot [
				domain=-0:1, 
				samples=100, 
				color=blue,
				]
				{exp(-x*10)}node[pos=0.43,anchor=west,fill=white,draw=black]{$\delta=\frac{1}{10}$};

				\addplot [
				domain=-0:1, 
				samples=100, 
				color=blue,
				]
				{exp(-x*20)}node[pos=0.5,anchor=west,fill=white,draw=black]{$\delta=\frac{1}{20}$};

				\end{axis}
				\end{tikzpicture}
				&
				\begin{tikzpicture}
				\begin{axis}[
				minor y tick num=1,
				minor x tick num=1,
				extra y ticks={0},
				extra x ticks={0},
				extra tick style={grid style={black},xticklabel=\empty},
				grid=both,
				axis lines = left,
				xlabel = $f(\textbf{g}_i)$,
				ylabel = {$Pc_i$(\textit{crossover} probability of the $i$th genes)},
				]
				%Below the red parabola is defined
				%Here the blue parabloa is defined
				\addplot [
				domain=-0:1, 
				samples=100, 
				color=blue,
				]
				{x^2}node[pos=0.6,anchor=east,fill=white,draw=black]{$\gamma=2$};
				\addplot [
				domain=-0:1, 
				samples=100, 
				color=blue,
				]
				{x^3}node[pos=0.55,anchor=east,fill=white,draw=black]{$\gamma=3$};
				\addplot [
				domain=-0:1, 
				samples=100, 
				color=blue,
				]
				{x^5}node[pos=0.50,anchor=east,fill=white,draw=black]{$\gamma=5$};
				\addplot [
				domain=-0:1, 
				samples=100, 
				color=blue,
				]
				{x^10}node[pos=0.45,anchor=east,fill=white,draw=black]{$\gamma=10$};		
				\end{axis}
				\end{tikzpicture}
				
			\end{tabular}
		}
	\end{center}
	
	\caption{Effects of parameter changes in the calculation of \textit{mutation} and \textit{crossover} probabilities.}
	\label{fig:curv}
	%\label{fig:onecol}
\end{figure*}
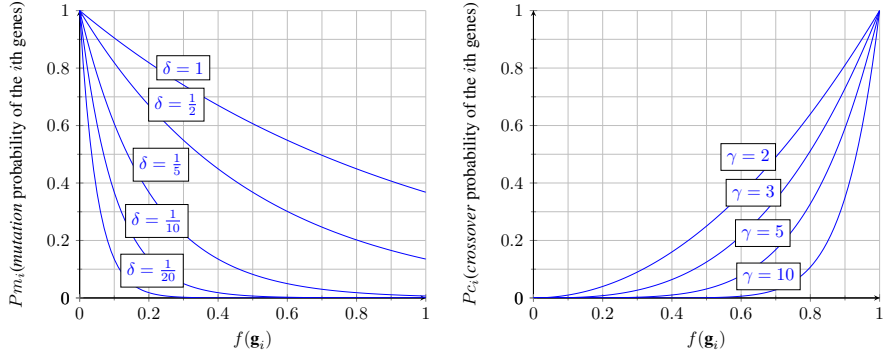

Where $Pc_i$ is \textit{crossover} probability of the $i$th genes, $f(\textbf{g}_i)$ is the fitness value for the $i$th gene, $norm()$ in the normalized function, and $ n $ is the power factor of the control function. The curves of different power factors based on normalized fitness represented on Figure \ref{fig:curv}.

The selection probability for \textit{mutation} operator is defined differently. As the fitness of a chromosome is lower, the probability of the \textit{mutation} should be higher. Therefore, an exponential function is used after fitness normalization and the complete formulation is given as follows:

\begin{equation} \label{eq:3}
Pm_i= exp\left(-\dfrac{norm\left(f\left(\textbf{g}_i\right)\right)}{\delta}\right)
\end{equation} 

Where $Pm_i$ is \textit{mutation} probability of the $i$th genes, and $ n $ is the power factor of the controlled Gaussian function.

\subsection{The Complete Algorithm}
The Complete algorithm of the proposed method is presented as following:\\

\begin{algorithm}
	\caption{The complete Novel Adaptive Genetic Algorithm Sample Consensus.}
	\begin{algorithmic}[1]
		\State {generate the first population randomly}
		\State {initialized replacement selection probability equally for all genes}
		\While{termination criteria is False}
		\State {Calculate the fitness of the population}
		\State {Calculate the probability of \textit{crossover} and \textit{mutation} for each chromosome base on adaptive formula}
		\State {Perform \textit{crossover} and \textit{mutation} operator base on their probability for each chromosome}
		\State {Roulette wheel training: increase the probability of the genes that involved in the selected parent}
		\EndWhile
	\end{algorithmic}
\end{algorithm}

\section{Experimental Results}

The evaluation of different RANSAC technique has been exhaustively studied in the literature. Here we designed two experiments to examine the robustness of our proposed algorithm to different rates of outliers against GASAC and RANSAC.

In the first experiment, the features of two consequent images are extracted and matched using SIFT algorithm. Since the feature matching process is not accurate enough, many mismatches occur. We discriminate the outlier and inlier matches manually and use them as ground-truth data.

RANSAC, GASAC and our method have a random nature. To have a fair comparison, we test each algorithm a hundred times and report the average results. On the other hand, RANSAC generates only one model in each generation while GASAC and our algorithm create a population of models in each generation. Therefore instead of the number of iterations, the results are reported per the number of generated models. 

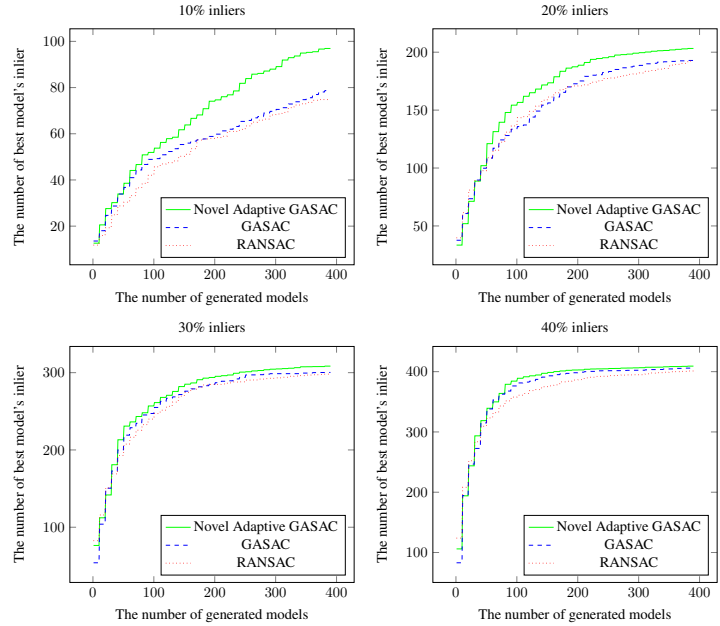
\begin{figure*}
	\begin{center}
		\resizebox{.8\columnwidth}{!}{%
			\begin{tabular}{c c}
				\begin{tikzpicture}
				\begin{axis}[
				title={10\% inliers},
				xlabel={The number of generated models},
				ylabel={The number of best model's inlier},
				legend pos=south east,
				]
				\addplot[color=green] table [x=X, y=W]{EVEx1plot00.dat};
				\addplot[color=blue,dashed] table [x=X, y=Y]{EVEx1plot00.dat};
				\addplot[color=red,dotted] table [x=X, y=Z]{EVEx1plot00.dat};
				\legend{Novel Adaptive GASAC,GASAC,RANSAC}
				\end{axis}
				\end{tikzpicture}
				&
				\begin{tikzpicture}
				\begin{axis}[
				title={20\% inliers},
				xlabel={The number of generated models},
				ylabel={The number of best model's inlier},
				legend pos=south east,
				]
				\addplot[color=green] table [x=X, y=W]{EVEx1plot01.dat};
				\addplot[color=blue,dashed] table [x=X, y=Y]{EVEx1plot01.dat};
				\addplot[color=red,dotted] table [x=X, y=Z]{EVEx1plot01.dat};
				\legend{Novel Adaptive GASAC,GASAC,RANSAC}
				\end{axis}
				\end{tikzpicture}
				\\
				\begin{tikzpicture}
				\begin{axis}[
				title={30\% inliers},
				xlabel={The number of generated models},
				ylabel={The number of best model's inlier},
				legend pos=south east,
				]
				\addplot[color=green] table [x=X, y=W]{EVEx1plot02.dat};
				\addplot[color=blue,dashed] table [x=X, y=Y]{EVEx1plot02.dat};
				\addplot[color=red,dotted] table [x=X, y=Z]{EVEx1plot02.dat};
				\legend{Novel Adaptive GASAC,GASAC,RANSAC}
				\end{axis}
				\end{tikzpicture}
				&
				\begin{tikzpicture}
				\begin{axis}[
				title={40\% inliers},
				xlabel={The number of generated models},
				ylabel={The number of best model's inlier},
				legend pos=south east,
				]
				\addplot[color=green] table [x=X, y=W]{EVEx1plot03.dat};
				\addplot[color=blue,dashed] table [x=X, y=Y]{EVEx1plot03.dat};
				\addplot[color=red,dotted] table [x=X, y=Z]{EVEx1plot03.dat};
				\legend{Novel Adaptive GASAC,GASAC,RANSAC}
				\end{axis}
				\end{tikzpicture}
				\\
				
			\end{tabular}
		}
	\end{center}
	\caption{The average number of inliers within the best model based on iterations of the algorithms in the first experiment. Each result is the average of a hundred runs. }
	\label{fig:ex1}
	%\label{fig:onecol}
\end{figure*}

Figure \ref{fig:ex1} depicts the average results of each algorithm based on different rates of inliers. The number of the generated model is fixed on 400 to prevent algorithms from early termination. The proposed method reach a better result in the presence of 40\% inliers in terms of both the number of detected inliers and the number of generated models to find the optimum.
In the presence of low rates of inliers, our proposed algorithm drastically outperforms the other two algorithms.

Using the adaptive \textit{selection} operator, the proposed method treats each generated model carefully so that the chance of losing a good individual is too small and bad ones are mutated with a high probability. Consequently, the model adapts the ratio of exploration based on its population; a good population would go through more exploitation and a bad one leads to an intense exploration of the solution area. 

The proposed \textit{mutation} also keeps the proper balance between exploration and exploitation. The roulette wheel is random in the first generations and explores thoroughly. Bad populations would not add much information to the roulette wheel since they do not have consensus in the same genes. The learning starts with the appearance of good individuals and gradually exploration is replaced with exploitation. The results of 10\% inliers confirm this claim. It shows that our proposed method explores better than GASAC and eventually finds good individuals. In comparison to RANSAC, our method deeply exploits the suitable areas and keeps the best ones for reproduction.

\begin{figure*}
	\begin{center}
		\resizebox{.5\columnwidth}{!}{%
			\begin{tikzpicture}
			\begin{axis}[
			xlabel={The number of generated models},
			ylabel={The number of best model's inlier},
			legend pos=south east,
			]
			\addplot[color=green] table [x=X, y=W]{EVplotAverage.dat};
			\addplot[color=blue,dashed] table [x=X, y=Y]{EVplotAverage.dat};
			\addplot[color=red,dotted] table [x=X, y=Z]{EVplotAverage.dat};
			\legend{Novel Adaptive GASAC,GASAC,RANSAC}
			\end{axis}
			\end{tikzpicture}	
		}
	\end{center}
	\caption{ The average number of inliers within the best model based on iterations of the algorithms in the second experiment which is the average of more than 27 thousand runs.}
	\label{fig:ex2}
	%\label{fig:onecol}
\end{figure*}
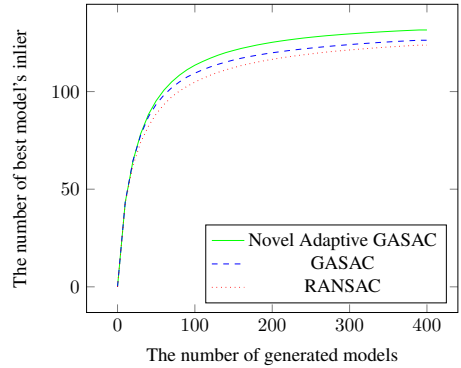
In the second test, all methods are applied to solve a stereo pose estimation problem based on the perspective-three-point technique \cite{gao2003complete}. The KITTI dataset \cite{geiger2012we} is used which has 21 different paths. Similar to the previous experiment, the number of generated models is fixed and same matched features are fed to all three algorithms. Generally, the inlier ratio is near to 40\%. 

The aim of this experiment is to evaluate the proposed method in a real-world problem in comparison with RANSAC and GASAC. Therefore, the algorithms are applied to all the frames of all paths in the KITTI dataset amounted to more than 27 thousands images. Finally, the average results of all models generated by three algorithms all paths are reported in Figure \ref{fig:ex2}. Each path contains more than thousand frames and averaging results of these numerous experiments guarantee a comprehensive and accurate conclusion.

Table \ref{table:tbex2} shows the average final score, number of found inliers, of best results for all runs. The higher score of our proposed method shows that our algorithm produces individuals with high fitness more frequently than RANSAC and GASAC. Note that the larger the number of found inliers, the more accurate the final model is.

\begin{table*}
	\caption{Average final score (the number of inlier within the best found model) for three methods in KITTI dataset.}
	\begin{center}
		\resizebox{\columnwidth}{!}{%	
			\begin{tabular}{ |c| c| c| c |c |c| }
				\hline
				
				& RANSAC
				& \multicolumn{2}{c}{GASAC} 
				& \multicolumn{2}{|c|}{Novel Adaptive GASAC} \\
				\cline{2-6}
				Methods: & results & results & Improvement & results & Improvement \\
				\hline\hline
				Average final score 	&	124	&	126		&	1.6	\%	&	132	&	6.4 \%	\\

				\hline
			\end{tabular}
		}
	\end{center}
	\label{table:tbex2}
	
\end{table*}

\section{Conclusion}
In real-time applications, it is crucial to have a proper balance between exploration and exploitation phases of the optimization algorithm. It would be more severe when there is a vast number of outliers in the dataset. In this paper, a new genetic algorithm sample consensus is proposed which adaptively manages the two phases. It provides a new adaptive strategy which successfully copes with various rates of outliers without any parameter adaptation. 
The proposed method utilizes a new adaptive \textbackslash\{\}textit\{selection\} operator which calculates the selection probabilities for each chromosome based on its normalized fitness. The algorithm strategy is to choose suitable chromosomes for \textbackslash\{\}textit\{crossover\} and mutate bad ones in each generation. Moreover, a learning roulette wheel is used to select replacement genes in \textbackslash\{\}textit\{mutation\}.
The proposed \textbackslash\{\}textit\{mutation\} starts fully random and the solution area is highly explored. Then it gradually learns the best replacements and carefully exploits the area. 
The proposed algorithm is first evaluated in the common feature matching problem and then applied to the stereo visual odometry dataset. The enormous number of runs in the provided experiments authorizes us to claim that our proposed method has higher robustness in different rates of outliers in addition to the higher number of found inliers.

\bibliography{evsac}

\end{document}